\documentclass[runningheads]{llncs}
\usepackage{graphicx}

\usepackage[dvipsnames]{xcolor}

\usepackage{booktabs}
\usepackage{tikz}
\usepackage{comment}
\usepackage{amsmath,amssymb} %
\usepackage{color}
\usepackage{multirow}

\usepackage{caption}

\usepackage{tabularx}

\DeclareMathOperator*{\argmin}{arg\,min}

\def\etal{\emph{et al.} }

\newcommand\eatpunct[1]{}

\newcommand{\KINETICSBOX}{Kinetics\textsuperscript{Box}}

\newcolumntype{Y}{>{\centering\arraybackslash}X}
\newcolumntype{P}[1]{>{\centering\arraybackslash}p{#1}}

\usepackage[accsupp]{axessibility}  %

\begin{document}
\pagestyle{headings}
\mainmatter
\def\ECCVSubNumber{2853}  %

\title{Beyond Transfer Learning: \\ Co-finetuning for Action Localisation}

\titlerunning{Beyond Transfer Learning: Co-finetuning for Action Localisation} 
\authorrunning{Arnab$ ^{\star}$, Xiong$ ^{\star}$ \etal} 
\author{Anurag Arnab\thanks{Equal contribution} \ Xuehan Xiong$^{\star}$ \ Alexey Gritsenko \ Rob Romijnders \ Josip Djolonga \ Mostafa Dehghani \ Chen Sun \ Mario Lučić\thanks{Equal advising} \ Cordelia Schmid$^{\star\star}$}
\institute{Google Research}

\maketitle

\begin{abstract}

Transfer learning is the predominant paradigm for training deep networks on small target datasets. %
Models are typically pretrained on large ``upstream'' datasets for classification, as such labels are easy to collect, and then finetuned on ``downstream'' tasks such as action localisation, which are smaller due to their finer-grained annotations.

In this paper, we question this approach, and propose co-finetuning --- simultaneously training a single model on multiple ``upstream'' and ``downstream'' tasks.
We demonstrate that co-finetuning outperforms traditional transfer learning when using the same total amount of data, and also show how we can easily extend our approach to multiple ``upstream'' datasets to further improve performance.
In particular, co-finetuning significantly improves the performance on rare classes %
in our downstream task, as it has a regularising effect, and enables the network to learn feature representations that transfer between different datasets.  %
Finally, we observe how co-finetuning with public, video classification datasets, we are able to achieve state-of-the-art results for spatio-temporal action localisation on the challenging AVA and AVA-Kinetics datasets, outperforming recent works which develop intricate models.

\end{abstract}

\section{Introduction}

\begin{figure}[t]
    \centering
    \includegraphics[width=1.0\textwidth]{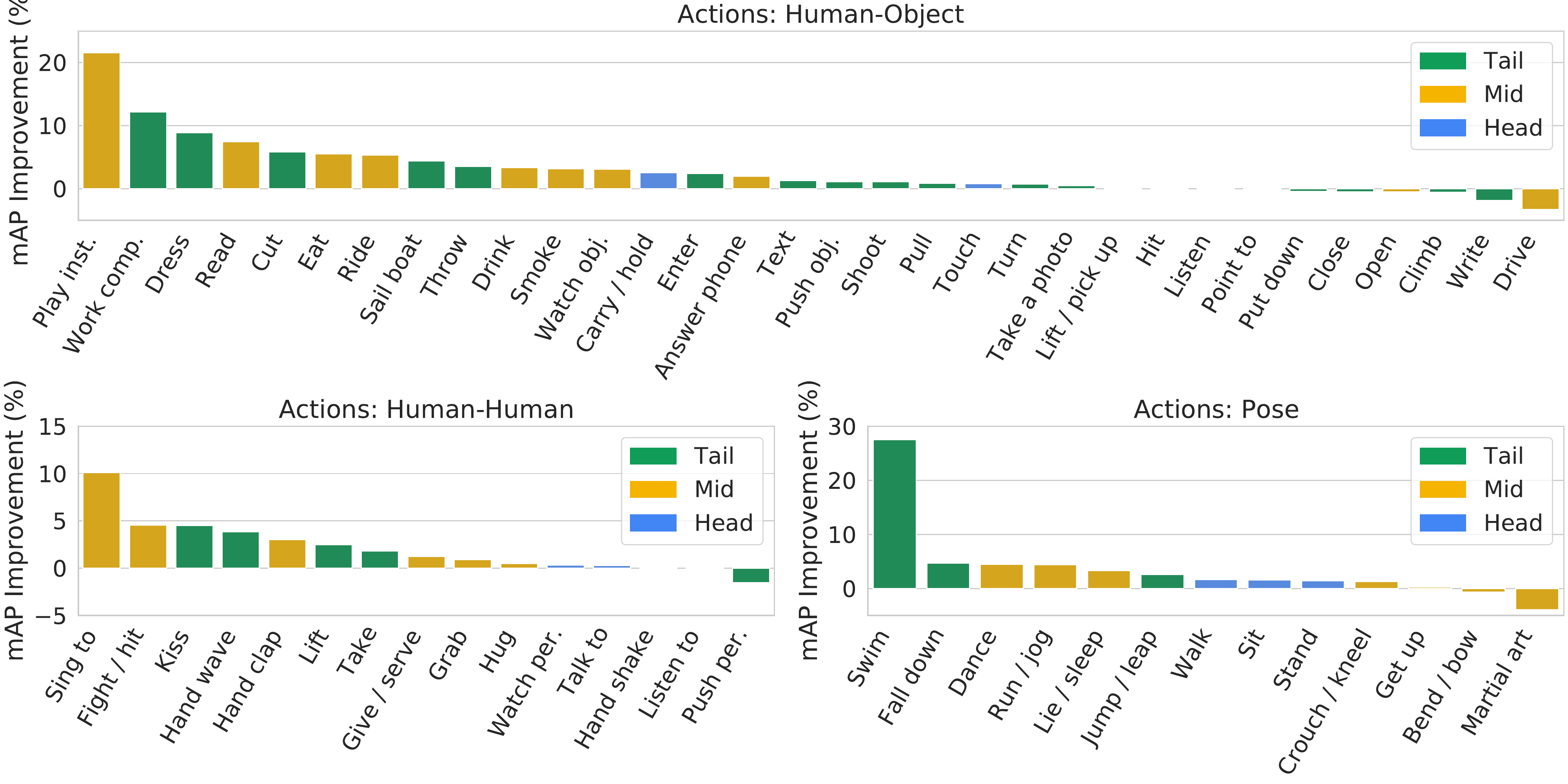}
    \caption{
    Improvements achieved by our co-finetuning strategy on the AVA dataset~\cite{gu_cvpr_2018}.
    AVA is a long-tailed dataset, and we attain significant improvements, particularly on the rare classes (shown by the ``Tail'' and ``Mid'' subsets) in the dataset.
    Our improvements are also consistent across the three action types defined in the AVA dataset.
    ``Head'', ``Mid'' and ``Tail'' classes are defined in Sec.~\ref{sec:experiments_ablation}.
    }
    \label{fig:teaser}
\end{figure}

The computer vision community has made impressive progress in video classification with deep learning, first with Convolutional Neural Networks (CNNs)~\cite{krizhevsky_neurips_2012,carreira_cvpr_2017,feichtenhofer_iccv_2019}, and more recently with transformers~\cite{vaswani_neurips_2017,arnab2021vivit,fan2021multiscale}.
However, progress in other more challenging  video understanding tasks, such as spatio-temporal action localisation~\cite{pan2021actor,tang2020asynchronous,arnab2021unified}, has lagged behind significantly in comparison.

One major reason for this situation is the lack of data with fine-grained annotations which are often not available for such complex tasks.
To cope with this challenge, the de facto approach adopted by most state-of-the-art approaches is transfer learning (popularised by~\cite{girshick2014rich}).
In the conventional setting, a model is first pre-trained on a large ``upstream'' dataset, which is typically labelled with classification annotations as they are less expensive
to collect.
The model is then ``finetuned'' on a smaller dataset, often for a different task, where fewer labelled examples are available~\cite{mensink2021factors}.
The intuition is that a model pre-trained on an auxiliary, ``upstream'' dataset learns generalisable features, and therefore its parameters do not need to be significantly updated during finetuning.

For video understanding, the most common ``upstream'' dataset is Kinetics~\cite{kay_arxiv_2017}, demonstrated by the fact that the majority of recent work addressing the task of spatio-temporal action localisation pretrain on it~\cite{fan2021multiscale,feichtenhofer_iccv_2019,pan2021actor,wu_cvpr_2019}.
Similarly for image-level tasks, ImageNet~\cite{deng_cvpr_2009} is the most common ``upstream'' dataset.

Our objective in this paper is to train more accurate models for spatio-temporal action detection, and we do so by proposing an alternate training strategy of co-finetuning. %
Instead of using the additional classification data in a separate pre-training phase, we simultaneously train for both classification and detection tasks.
Intuitively, the additional co-finetuning datasets can act as a regulariser during training, benefiting in particular the rare classes in the target dataset which the network could otherwise overfit on. %
Moreover, discriminative features which are learned for classification datasets may also transfer to the detection dataset, even though the target task and labels are different.

Our thorough experimental analyses confirm these intuitions.
As shown by Fig.~\ref{fig:teaser}, our co-finetuning strategy improves spatio-temporal action localisation results for the vast majority of the action classes on the AVA dataset, improving substantialy on the rarer classes with few labelled examples. %
In particular, co-finetuning performs better than traditional transfer learning when using the same total amount of data. %
Moreover, with co-finetuning we can easily make use of additional ``upstream'' classification datasets during co-finetuning to improve results even further. %

Our approach is thus in stark contrast to previous works on spatio-temporal action detection which develop complex architectures to model long-range relationships using graph neural networks~\cite{arnab2021unified,wang_eccv_2018,sun_eccv_2018,baradel_eccv_2018,zhang_tokmakov_cvpr_2019}, external memories~\cite{wu_cvpr_2019,pan2021actor,tang2020asynchronous} and additional object detection proposals~\cite{wang_eccv_2018,baradel_eccv_2018,tang2020asynchronous,zhang_tokmakov_cvpr_2019}.
Instead, we use a simple detection architecture and modify only the training strategy to achieve higher accuracies, outperforming prior work.

We conduct thorough ablation analyses to validate our method, and make other findings too, such as the fact that although Kinetics~\cite{kay_arxiv_2017} is the most common upstream pretraining dataset for video tasks, and used by all previous work addressing spatio-temporal action detection on AVA, Moments in Time~\cite{monfort_pami_2019} is actually better.
Finally, a by-product of our strategy of co-finetuning with classification tasks is that our network can simultaneously perform these tasks, and is competitive with the state-of-the-art on these datasets too.

\section{Related Work}

We first discuss transfer- and multi-task learning, which our proposed co-finetuning approach is related to.
We then review action detection models, as this is our final task of interest.

\paragraph{Transfer- and multi-task learning}
The predominant paradigm for training a deep neural network is to pre-train it on a large ``upstream'' dataset with many annotations, and then to finetune it on the final ``downstream'' dataset of interest.
This approach was notably employed by the R-CNN detector~\cite{girshick2014rich} which leveraged pre-trained image classification models on ImageNet~\cite{deng_cvpr_2009}, for object detection on the significantly smaller Pascal VOC detection dataset~\cite{everingham2015pascal}.
Since then, this strategy has become ubiquitous in training deep neural networks (which require large amounts of data to fit) across a wide range of computer vision tasks~\cite{mensink2021factors}.
The training strategy of Girshick~\etal\cite{girshick2014rich} is a form of ``inductive transfer learning'' based on the taxonomy of~\cite{pan2009survey}, and we simply refer to it as ``traditional transfer learning'' in this paper.

We further note that it is common to perform multiple stages of transfer, in which case the ``downstream'' task can become the ``upstream'' task for the next stage of training~\cite{mensink2021factors}.
For example, video classification models are often initialised from image models pretrained on ImageNet-1K or ImageNet-21K~\cite{deng_cvpr_2009}, and then finetuned on Kinetics~\cite{wang2016temporal,carreira_cvpr_2017,arnab2021vivit,bertasius_arxiv_2021,liu2022video,zhang2021vidtr}.
These video classification models are then again finetuned on the final task of interest, such as spatio-temporal action detection on AVA~\cite{girdhar_cvpr_2019,wang_cvpr_2018,zhao2021tuber}.
In this work, we focus on co-finetuning video classification and action detection datasets together.

Our co-finetuning strategy, of jointly training a model on multiple upstream and downstream datasets simultaneously, is closely related to multi-task learning.
Multi-task learning aims to develop models that can address multiple tasks whilst sharing parameters and computation among them~\cite{caruana1997multitask}.
However, in computer vision, the focus of many prior works has been to predict multiple outputs (for example semantic segmentation and surface normals)~\cite{eigen2015predicting,kokkinos2017ubernet,kendall2018multi,kurin2022defense} given a single input image.
These works have typically observed that although these multi-task models are more versatile, their accuracies are lower than single-task models.
Moreover, this accuracy deficit worsens as the number of tasks grows, or if multiple unrelated tasks are simultaneously performed~\cite{kokkinos2017ubernet,zamir2018taskonomy,sener2018multi}.
Training a model to perform multiple tasks on the same input often results in one task dominating the other, and numerous strategies have been proposed to mitigate this problem by adapting the losses or gradients from each task~\cite{sener2018multi,kendall2018multi,chen2020just}.

Our scenario differs, in that although our network is capable of performing multiple tasks, it only performs a single task at a time for a given input which is also how multi-task learning is typically referred to in natural language processing~\cite{collobert2008unified,tay2020hypergrid,raffel2019exploring}.
However, this setting is not common in vision, with Maninis~\etal\cite{maninis_cvpr_2019} referring to it as ``single tasking of multiple tasks.''

Our decision of performing a single task at a time ensures that our training strategy is simple, and does not require additional hyperparameters to stabilise, like previous multi-task learning methods in computer vision.
Moreover, we are primarily interested in improving performance on a target task, and using the additional datasets to learn more generalisable features and for regularisation.

\paragraph{Spatio-temporal action detection models} 
Current state-of-the-art action detection models~\cite{arnab2021unified,pan2021actor,fan2021multiscale,tang2020asynchronous} are based on the Fast R-CNN architecture~\cite{girshick_iccv_2015}, and use external person detections as region proposals which are pooled and classified.
We note, however, that some recent works~\cite{chen2021watch,zhao2021tuber} build off DETR~\cite{carion_eccv_2020} and do not require external proposals, although they are not as performant.

Localising actions in space and time often requires capturing long-range contextual information, and state-of-the-art approaches have explicitly modelled this context to improve results.
A common theme is to use graph neural networks to explicitly model relationships between actors, objects and the scene~\cite{arnab2021unified,baradel_eccv_2018,sun_eccv_2018,wang_eccv_2018}.
Many approaches also use off-the-shelf object detectors to locate relevant scene context~\cite{baradel_eccv_2018,herzig2021object,wang_eccv_2018,zhang_tokmakov_cvpr_2019}.
To model long-range temporal contexts spanning several minutes,~\cite{wu_cvpr_2019,pan2021actor,tang2020asynchronous} have proposed external memory banks which are populated by first running a video feature extractor over the input video, and are thus not trained end-to-end.
The most accurate methods have included a combination of external memories and spatio-temporal graphs~\cite{pan2021actor,tang2020asynchronous}, and are thus complex models which require complex training and inference procedures due to the external memory bank.

Our proposed approach, in stark contrast, uses a simple Fast R-CNN architecture, without any explicit graph modelling~\cite{arnab2021unified,zhang_tokmakov_cvpr_2019,sun_eccv_2018}, auxillary object detections~\cite{herzig2021object,wang_arxiv_2020,zhang_tokmakov_cvpr_2019} or external memories~\cite{wu_cvpr_2019,pan2021actor,tang2020asynchronous}, and achieves improvements over the baseline model that are competitive or even better than the complex models of~\cite{arnab2021unified,pan2021actor,sun_eccv_2018,wu_cvpr_2019}, by simply changing the training strategy.

\section{Proposed Approach}

\begin{figure}[t]
    \centering
    \includegraphics[width=0.8\textwidth]{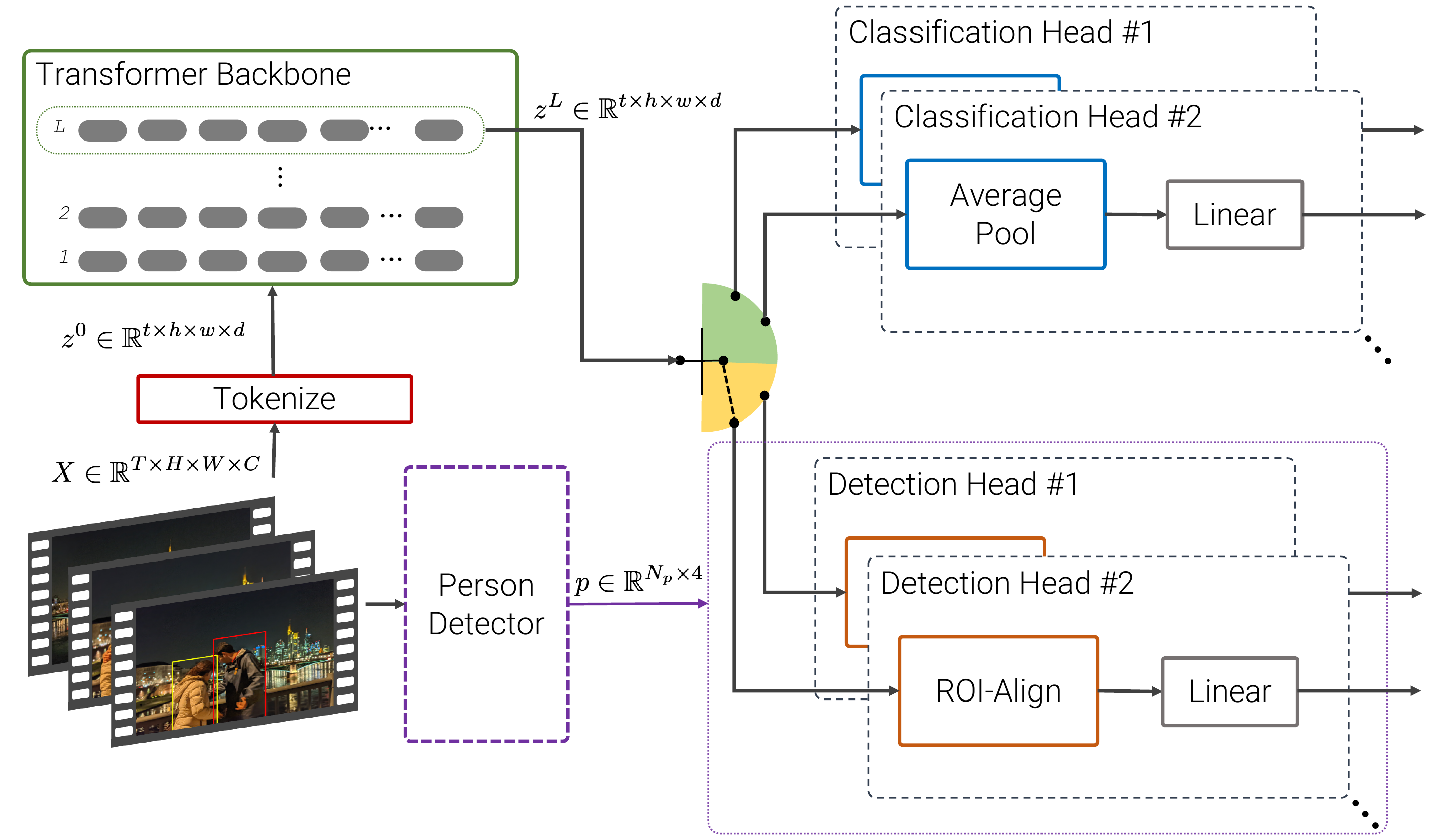}
    \caption{Overview of our model architecture.
    A transformer backbone is used to extract spatio-temporal features which are then processed by task-specific heads.
    Note that we co-finetune on multiple tasks simultaneously.
    }
    \label{fig:model}
\end{figure}

Our objective is to train a model for spatio-temporal detection, using additional video classification data to improve our model's performance.
We first describe our model in Sec.~\ref{sec:method_model}, before detailing how we co-finetune it using additional classification data in Sec.~\ref{sec:method_cotraining}.
Finally, we discuss our training strategy in Sec.~\ref{sec:method_discussion}.

\subsection{Model}
\label{sec:method_model}

Our main objective is to study co-finetuning strategies for training action detection models. As such, we choose a transformer encoder model with separate heads for each classification and detection dataset that we co-finetune with (Fig.~\ref{fig:model}).

Given an input $\mathbf{x} \in \mathbb{R}^{T \times H \times W \times C}$, we first extract $N$ tokens from the input video, $\mathbf{z}^{0} = \{z_1, z_2, \ldots, z_N\}$, with $\mathbf{z} \in \mathbb{R}^{t \times h \times w \times d}$. 
Here, $d$ denotes the hidden dimension of the model, and the spatio-temporal dimensions of the tokens, ($t, h, w$), depend on the tubelet size~\cite{arnab2021vivit} when tokenizing the input. %
These tokens are then processed by a transformer encoder consisting of $L$ layers to result in $\mathbf{z}^L$.
Depending on the task (video classification, or action detection), the encoded tokens, $\mathbf{z}^L$, are forwarded through a dataset-specific head, as shown in Fig.~\ref{fig:model}.

\subsubsection{Detection head}
As shown in Fig.~\ref{fig:model}, our detection head follows the Fast R-CNN meta-architecture~\cite{girshick_iccv_2015}.
We adopt this method as current and previous state-of-the-art approaches~\cite{pan2021actor,feichtenhofer_iccv_2019,wu_cvpr_2019,fan2021multiscale} have used it.
Specifically, we use person detections, $\mathbf{p} \in \mathbb{R}^{N_p \times 4}$, obtained from an off-the-shelf person detector, to spatio-temporally pool the encoded tokens, $\mathbf{z}^L$, using ROI-Align~\cite{he_iccv_2017}.
Here, $N_p$ denotes the number of people detected for the particular input video.
We then obtain our predicted logits for the $j^{th}$ person-proposal as $\mathbf{y}_j$ using a linear projection.
This is described by
\begin{align}
    \mathbf{z}^{p}_j &= \text{ROI-Align}(\mathbf{z}^L, p_j), \\
    \mathbf{y}_j &= \mathbf{W}\mathbf{z}^{p}_j.
\end{align}
Note that our person detections, $\mathbf{p}$, are for the central keyframe of the video clip following~\cite{pan2021actor,feichtenhofer_iccv_2019,wu_cvpr_2019,fan2021multiscale}.
For ROI-pooling, we extend the bounding box temporally to form a ``tube'' following prior work.

\subsubsection{Classification head}
The video classification head is a simple linear classifier.
We extract a representation of the video by spatio-temporally %
averaging all the encoded tokens.
Therefore, $\mathbf{z}^{v} = \text{Average}(\mathbf{z}^L) \in \mathbb{R}^d$, and the logits predicted for each class for the $i^{th}$ dataset are $\mathbf{y} = \mathbf{W}^{i}\mathbf{z}^{v}$.

\subsection{Co-finetuning Strategy}
\label{sec:method_cotraining}

The de-facto approach for training deep architectures for computer vision tasks is to pre-train the model on a large ``upstream'' dataset, such as ImageNet~\cite{deng_cvpr_2009}, and then to finetune it on the smaller target dataset.
Each training procedure (pre-training or finetuning), can be denoted by
\begin{equation}
    \argmin_{\theta} \mathop{\mathbb{E}}_{(x, y) \in \mathcal{D}} L(f(\mathbf{x}; \theta), \mathbf{y}),
    \label{eq:training}
\end{equation}
where $\mathcal{D}$ denotes the dataset consisting of pairs of videos, $\mathbf{x}$ and ground truth labels, $\mathbf{y}$, $L$ is the cross-entropy loss function, and $f$ is a neural network with model parameters, $\theta$.
If we denote $\theta = \{\theta_{b}, \theta_{h}\}$ as the parameters of the backbone and head respectively, ``pre-training'' can be understood as optimising all model parameters from random initialisation, %
whereas ``finetuning'' initialises the backbone model parameters, $\theta_b$ from pre-training, and trains new head parameters, $\theta_h$, for the final task of interest.
Initialisation of network parameters $\theta$ is critical, as it is a non-convex optimisation problem.

Note that it is common, particularly in video understanding, to include multiple stages of pre-training.
For example, many video models are initialised from networks pretrained on ImageNet, then finetuned on Kinetics and then finally finetuned for spatio-temporal detection~\cite{girdhar_cvpr_2019,kopuklu2019you,wu_cvpr_2019,zhao2021tuber}.

In our proposed approach, we simultaneously finetune our model on multiple datasets:
This includes our target dataset of interest, %
as well as other datasets from which our model can learn expressive features, and which may help to regularise the model.
We can express this as
\begin{equation}
    \argmin_{\theta} \mathop{\mathbb{E}}_{(x, y) \in \mathcal{D}_1 \cup \mathcal{D}_2 \ldots \cup \mathcal{D}_{N_d}}  L(f(\mathbf{x}; \theta), \mathbf{y}),
    \label{eq:cotraining}
\end{equation}
where $N_d$ denotes the total number of datasets.
Note that our model parameters are now $\theta = \{\theta_{b}, \theta_{h_1}, \theta_{h_2}, \ldots, \theta_{h_{N_d}}\}$ as we have separate head parameters, $\theta_{h_i}$, for each of our $N_d$ co-finetuning datasets.

\subsubsection{Implementation}

To co-finetune on multiple datasets simultaneously as in Eq.~\ref{eq:cotraining}, we construct each training minibatch by sampling examples from one particular dataset.
We then perform a parameter-update using SGD with gradients computed from this batch, before sampling another dataset and iterating.
We consider the following two alternatives for sampling minibatches:

\paragraph{Alternating} Given $N_d$ datasets, we sample batches from datasets sequentially.
For example, we first sample a batch from dataset $\mathcal{D}_1$, then $\mathcal{D}_2, \ldots \mathcal{D}_{N_d}$ before continuing from $\mathcal{D}_1$ again.

\paragraph{Weighted sampling} We sample batches from a dataset with a probability proportional to the number of examples in the dataset.
This corresponds to concatenating several datasets together, and drawing batches randomly.
It is also possible to set the sampling weights for each dataset, %
but we avoid this approach to avoid introducing additional hyperparameters to tune.

\subsubsection{Discussion}

Note that it is also possible to construct a minibatch using a mixture of examples from each dataset.
However, we found this approach to be less computationally efficient in initial experiments and did not pursue it further.

Finally, if we constructed our batches such that we first sampled only examples from dataset $\mathcal{D}_1$, then $\mathcal{D}_2$ and so on, this would be equivalent to finetuning on each dataset sequentially, with the difference that the optimiser state (such as first- and second-order moments of the gradient~\cite{qian1999momentum,kingma2014adam}) would be carried over from one dataset to the next, which is not typically done when finetuning.

\subsection{Intuition behind co-finetuning}
\label{sec:method_discussion}

In this section, we explain our intuitions behind why co-finetuning could be effective, given that by sequentially finetuning a model on $N_d$ different datasets, the model would have access to the same total amount of data.

By co-finetuning on multiple datasets simultaneously, it is possible for the model to learn general, discriminitive visual patterns on one dataset that transfer to the other.
And this effect may be more pronounced when the target dataset (such as AVA~\cite{gu_cvpr_2018} for spatio-temporal action detection) is small and contains many tail classes with few labelled examples.
Sequentially finetuning a model on multiple datasets may not have the same property due to ``catastrophic forgetting''~\cite{french1999catastrophic,kirkpatrick2017overcoming} --- the propensity of neural networks to degrade at previous tasks when finetuned on a new one.

Another reason is that deep neural networks, and especially transformers, are prone to overfitting, as evidenced by the numerous regularisation techniques that are typically employed when training them~\cite{touvron_arxiv_2020,arnab2021vivit,bello2021revisiting,wightman2021resnet}.
Training with additional data, even if the additional co-finetuning datasets are not for the task of interest, can therefore remediate this problem by acting as a regulariser.
In particular, we would expect the benefits to be the most apparent on small and/or imbalanced datasets (since the network may otherwise easily memorise the classes with few examples).
Datasets for more complex video understanding tasks, such as action detection, are typically small due to their annotation cost.

We now validate these hypotheses experimentally in the next section.

\section{Experiments}

\subsection{Experimental Setup} %
\label{sec:experiments_setup}

\subsubsection{Datasets}

\textit{AVA}~\cite{gu_cvpr_2018} is the largest dataset for spatio-temporal action detection, consisting of 430, 15-minute video clips obtained from movies.
The dataset consists of 80 atomic actions, of which 60 are used for evaluation~\cite{gu_cvpr_2018}.
These actions are annotated for all actors in the video, where one person is typically simultaneously performing multiple actions.
The dataset annotates keyframes at every second in the video, and following standard practice, we report the Frame AP at an IoU threshold of 0.5 using the annotations from version 2.2~\cite{gu_cvpr_2018}.

\textit{Kinetics}~\cite{kay_arxiv_2017} is a collection of %
clip-classification datasets consisting of 10 second video clips focusing on human actions.
The Kinetics datasets have grown progressively, with Kinetics 400, 600 and 700, annotated with 400, 600 and 700 action classes respectively. The dataset sizes range from 240,000 (Kinetics 400) to 530,000 (Kinetics 700).
To avoid confusion with AVA-Kinetics, described below, we denote these datasets as Kinetics\textsuperscript{Clip}.

\textit{AVA-Kinetics}~\cite{li2020ava} adds detection annotations, following the AVA protocol, to a subset of Kinetics 700 videos. Note that only a single keyframe, in a 10-second Kinetics clip, is annotated.
To avoid confusion with Kinetics\textsuperscript{Clip}, we use Kinetics\textsuperscript{Box} to denote the Kinetics videos with detection annotations.
Therefore, the training and validation splits of AVA-Kinetics are the union of the respective AVA and Kinetics\textsuperscript{Box} dataset splits.

\textit{Moments in Time}~\cite{monfort_pami_2019} consists of 800,000 clips, each 3-seconds long and annotated with one of the 339 classes.
The videos involve people, animals, objects, or natural phenomena and capture the gist of a dynamic scene.

\textit{Something-Something v2}~\cite{goyal_iccv_2017} consists of about 220,000 short video clips annotated with 174 classes showing humans interacting with everyday objects.
In contrast to other datasets, similar objects and backgrounds appear in videos across different classes, meaning that a model must recognise fine-grained motion cues to distinguish different classes.

\subsubsection{Implementation details}

Our transformer backbone architecture is the ViViT Factorised Encoder~\cite{arnab2021vivit}, with $16\times16\times2$ tubelets, following the authors' public implementation~\cite{arnab2021vivit,dehghani2021scenic}. %
We chose this backbone as it is a state-of-the-art video classification model.
Note that ViViT, like other popular transformers for video~\cite{bertasius_arxiv_2021,liu2022video,zhang2021vidtr} is initialised from a ViT image model pretrained on ImageNet-21K~\cite{deng_cvpr_2009} when training for video tasks, and we follow this method unless otherwise stated.

We use the same person detections as prior work~\cite{wu_cvpr_2019,feichtenhofer_iccv_2019,fan2021multiscale,arnab2021unified}.
For AVA, these were obtained using a Faster R-CNN~\cite{ren_neurips_2015} detector, and released publicly by~\cite{wu_cvpr_2019}, achieving an AP of 93.9 for person detection on the AVA validation set.
For AVA-Kinetics, these were obtained using CenterNet~\cite{zhou2019objects} by~\cite{li2020ava}, obtaining an AP of 77.5 on AVA-Kinetics for person detection.
Furthermore, note that we report results from processing a single view for all tasks.

All models are trained on 32 frames with a temporal stride of 2.
We train our models for 15 epochs using synchronous SGD with momentum of 0.9, following a cosine learning rate schedule with linear warm-up for the first 2.5 epochs.
An epoch is defined using the sum of the size of each dataset.

\subsection{Ablation Study}
\label{sec:experiments_ablation}

We first conduct thorough ablation studies of our co-finetuning approach in Tables~\ref{tab:ablation_cotraining_same_data} through~\ref{tab:ablation_wts}.
For rapid experimentation, unless otherwise stated, we use a ViViT-Base Factorised Encoder as our backbone model, where videos are resized such that the longest side has a resolution of 224 (denoted as 224p).
We also use the ``Weighted'' minibatch sampling strategy (Sec.~\ref{sec:method_cotraining}) unless otherwise stated.

\subsubsection{Comparison of co-finetuning and traditional finetuning with the same data}
Tables~\ref{tab:ablation_cotraining_same_data} and~\ref{tab:ablation_cotraining_3_datasets} compare the effect of traditional transfer learning, and co-finetuning, with the same total amount of data, showing that co-finetuning achieves superior action detection results on AVA, as well as improved video classification accuracy, in all cases.

\begin{table}[t]
	\caption{Comparison of traditional finetuning, and our proposed co-finetuning, with the same total amount of data. 
	Observe how co-finetuning consistently outperforms traditional finetuning.
	We report single-view accuracies for all datasets, using a ViViT-Base backbone at 224p frame resolution.
	}
	\begin{minipage}{.48\linewidth}
		\centering
		(a) Kinetics 400 as upstream classification dataset
		\begin{tabularx}{\linewidth}{YcYY}
			\toprule
			Pretrain & Finetune & AVA & K400 \\
			\midrule
			K400 & -- & -- & 74.9  \\ %
			K400 & AVA  & 23.4 & -- \\ %
			\midrule
			-- & K400 + AVA & \textbf{25.2} & \textbf{76.2}  \\  %
			\bottomrule
	\end{tabularx}
	\end{minipage} \hfill
	\begin{minipage}{.48\linewidth}
		\centering
		(b) Moments in Time as upstream classification dataset
		\begin{tabularx}{\linewidth}{YcYY}
			\toprule
			Pretrain & Finetune & AVA & MiT\\
			\midrule
			MiT & -- & -- & 36.8 \\ %
			MiT & AVA & 24.8 & -- \\ %
			\midrule
			-- & MiT + AVA & \textbf{26.1} & \textbf{38.1}\\ %
			\bottomrule
	\end{tabularx}
	\end{minipage}
	\label{tab:ablation_cotraining_same_data}
\end{table}

\begin{table}[t]
	\centering
	\caption{Comparison of traditional finetuning, and our proposed co-finetuning, using two upstream datasets. Observe how co-finetuning consistently outperforms traditional finetuning with the same total amount of data, even when we account for the fact that there are multiple orders in which two upstream datasets can be pretrained in with traditional finetuning (K400$\to$MiT indicates that we first pretrain on K400 and then MiT).
	We report single-view accuracies for all datasets using a ViViT-Base model at 224p frame resolution.
	}
	\begin{tabular}{ccP{1cm}P{1cm}P{1cm}}
		\toprule
		Pretrain & Finetune & AVA & K400 & MiT \\
		\midrule
		K400 & MiT & -- & -- & 37.2 \\ %
		MiT & K400 & -- & 76.3 & -- \\ %
		K400$\to$MiT & AVA & 25.3 & -- & -- \\ %
		MiT$\to$K400 & AVA & 25.1 & -- & -- \\ %
		\midrule
		-- & MiT + K400 + AVA & \textbf{26.3} & \textbf{76.7} & \textbf{38.9} \\ %
		\bottomrule
	\end{tabular}
	\label{tab:ablation_cotraining_3_datasets}
\end{table}

Note that in all experiments, our ViViT backbone model is initialised from an ImageNet-21K pretrained checkpoint, following common practice in training transformer models for video~\cite{arnab2021vivit,bertasius_arxiv_2021,liu2022video,yan2022multiview,zhang2021vidtr}, and we omit this from the ``Pretrain'' column for clarity.

Specifically, Tab.~\ref{tab:ablation_cotraining_same_data}(a) shows that co-finetuning on Kinetics 400 and AVA jointly improves the mAP on AVA by 1.8 points, or 7.7\% relative, whilst using the same total amount of data.
An added advantage of co-finetuning is that the model can perform multiple tasks --- unlike traditional transfer learning, the model does not ``forget'' the upstream dataset~\cite{french1999catastrophic} as it is simultaneously optimised to perform both tasks.
We can see that although single-crop classification performance on Kinetics 400 increases by 1.3 points. 
Moreover, there is a similar increase for Moments in Time (MiT) in Tab.~\ref{tab:ablation_cotraining_same_data}(b).

Table~\ref{tab:ablation_cotraining_same_data}(b) shows that co-finetuning also improves AVA action localisation accuracy, achieving an improvement of 0.8 points.
Note that existing work reporting results on AVA~\cite{arnab2021unified,girdhar_cvpr_2019,feichtenhofer_iccv_2019,pan2021actor,tang2020asynchronous,wu_cvpr_2019} have used Kinetics as the upstream classification dataset for transfer learning.
However, Tab.~\ref{tab:ablation_cotraining_same_data}(b) shows that pretraining on MiT outperforms Kinetics 400 by 1.4 points.
Note that whilst MiT contains more video clips than Kinetics 400 (800,000 versus 240,000), its clips are also much shorter at 3 seconds compared to the 10 seconds of Kinetics, and hence the total duration of video in both datasets is similar. %

Finally, Tab.~\ref{tab:ablation_cotraining_3_datasets} considers the scenario where we have two upstream classification datasets, Kinetics 400 and MiT.
With traditional transfer learning, there are two orders in which we can pretrain on the upstream clip classification datasets: first on Kinetics 400 and then MiT, or vice versa.
We observe that co-finetuning outperforms both of these alternatives on AVA action detection, and moreover, does not require the additional training hyperparameter of choosing which order to use the upstream clip classification datasets.

\subsubsection{Adding more co-finetuning datasets}
Table~\ref{tab:cotrain_adding_more_datasets} shows how adding more clip-classification datasets for co-finetuning improves our spatio-temporal action detection performance on AVA.
We observe progressive improvements from adding Kinetics, Moments in Time (MiT) and Something-Something (SSv2) respectively, although the gains diminish with additional datasets.
We chose these clip classification datasets for co-finetuning as they are the largest available public datasets that we are aware of.
Note that Kinetics and Moments in Time consist of videos that are similar in domain to AVA as they are collected from YouTube, and feature people.
SSv2~\cite{goyal_iccv_2017}, on the other hand, consists of videos from a different domain, consisting mostly of objects being manipulated, and requires a model to capture fine-grained motion patterns.
Nevertheless, we observe noticeable improvements from using SSv2, suggesting that the additional data has a regularising effect on the model, and that motion cues learned from SSv2 are useful for AVA detection.

\begin{table}[t]
	\centering
	\caption{
		By using more clip-classification datasets for co-training, spatio-temporal action detection performance, measured by the mAP on AVA, gradually increases.
		Note how the improvement is primarily in the rare classes, denoted by the ``Mid'' and ``Tail'' categories.
		Refer to Sec.~\ref{sec:experiments_ablation} and Fig.~\ref{fig:per_class_results} for additional details on how these splits were constructed.
	}
	\begin{tabularx}{0.96\linewidth}{Xllll}
		\toprule
		Training / Co-training datasets & Head & Mid & Tail & Overall \\
		\midrule
		K400$\to$AVA traditional finetuning baseline&  65.5 & 29.7 & 6.9 & 23.4\\
		\midrule
		K400 + AVA 						  & 65.6 & 31.8 & 8.9 & 25.2 \\  %
		K400 + MiT + AVA 			 & 66.1 & 33.7 & 9.5 & 26.3 \\	 %
		K400 + MiT + SSv2 + AVA & $\textbf{66.7}_{\color{Green}+1.2}$ & $\textbf{34.1}_{\color{Green}+4.4}$ & $\textbf{10.1}_{\color{Green}+3.2}$ & $\textbf{26.8}_{\color{Green}+3.6}$ \\ %
		\bottomrule
	\end{tabularx}
	\label{tab:cotrain_adding_more_datasets}
\end{table}

\subsubsection{What classes does co-finetuning benefit?}
One of our hypotheses for co-finetuning was that it enables a model to learn discriminative visual patterns on one dataset that transfer to the other, and that this would particularly benefit rare classes in the target dataset that the model with otherwise overfit on.

To validate this, we split the 60 class labels from AVA into ``Head'', ``Mid'' and ``Tail'' classes.
``Head'' classes are defined as those with more than 10,000 labelled ground truth examples in the training set, whereas ``Tail'' classes have less than 1000 instances.
Classes which don't fall into either of these categories are defined as ``Mid'' classes.
There are 8 ``Head'', 23 ``Mid'' and 29 ``Tail'' classes respectively, as can be seen in Fig.~\ref{fig:teaser} and~\ref{fig:per_class_results}.

As shown in Tab.~\ref{tab:cotrain_adding_more_datasets}, adding more co-finetuning datasets improves our AVA detection mAP for ``Mid'' and ``Tail'' classes.
In particular, our best co-finetuned model improves ``Mid'' classes by 4.4 points, or 14.8\% relative.
The relative improvement for ``Tail'' classes is even more, at 46.4\%.
Figure~\ref{fig:teaser} details this improvement further, showing the difference with respect to the baseline by co-finetuning on each action class.

\subsubsection{Improvements from training strategy compared to modelling}
Table~\ref{tab:cotrain_adding_more_datasets} shows that by leveraging public clip classification datasets, we are able to increase performance on AVA, measured by the mAP, by 3.6 points.
To put this in context, we notice that the spatio-temporal actor-object graph proposed by~\cite{arnab2021unified} improved the author's baseline by 2.2 points.
The external memory of~\cite{wu_cvpr_2019} which requires precomputing video features for the entire video offline before training and inference obtains an improvement over the baseline of 2.9 points. 
The method of~\cite{pan2021actor}, which combines external memories and spatio-temporal graphs, improves over its respective baseline by 3.4 points, but is significantly more complex.
Therefore, we conclude that co-finetuning on public clip-classification datasets provides a simple method of increasing action detection models by amounts commensurate with, or even higher than, the latest state-of-the-art methods.
We caution, however, that definitive comparisons are difficult to make as each of the aforementioned methods uses slightly different baselines and code bases.

\begin{table}[t]
	\begin{minipage}{.48\linewidth}
		\centering
		\caption{Effect of different minibatch sampling strategies. We co-finetune on K400, MiT and AVA simultaneously with a ViViT-B 224p model.}
		\vspace{\baselineskip}
	\begin{tabular}{lccc}
		\toprule
		& AVA  & K400 & MiT  \\ \midrule
		Alternating 			   & 25.0 & 76.4 & 34.8 \\
		Weighted sampling    & \textbf{26.3} & \textbf{76.7} & \textbf{38.9} \\ \bottomrule
		\end{tabular}
	\label{tab:ablation_minibatch_sampling}
	\end{minipage} \hfill
	\begin{minipage}{.48\linewidth}
		\centering
		\caption{Effect of using the web-crawled WTS dataset~\cite{stroud2020learning} for initial pre-training, using a ViViT-B 512p model.
		Even with stronger initialisation, co-finetuning provides benefits.		
		}
		\begin{tabular}{lc}
			\toprule
			& AVA  \\ \midrule
			Traditional finetuning & 33.3 \\
			Co-finetuning          & \textbf{34.2} \\ \bottomrule
		\end{tabular}
		\label{tab:ablation_wts}
	\end{minipage}
\end{table}

\subsubsection{Minibatch sampling strategy for co-finetuning}
Table~\ref{tab:ablation_minibatch_sampling} evaluates the effect of the two different minibatch sampling strategies described in Sec.~\ref{sec:method_cotraining}.
The ``alternating'' minibatch strategy samples an equal number of minibatches from each of the datasets during co-finetuning, regardless of the size of the dataset.
Consequently, the accuracy on the largest dataset (MiT) decreases as not enough iterations of backpropagation are applied to it.
On the other hand, the accuracy on the smallest dataset (AVA) decreases too as too many optimisation iterations are performed on it, and the network begins to overfit on it.
The ``Weighted sampling'' strategy does not suffer these problems, and performs better overall, as batches are drawn from the co-finetuning datasets in proportion to the size of each dataset.
This experiment thus justifies our decision to use the ``Weighted sampling'' strategy in all of our experiments.

\subsubsection{Leveraging large-scale pretrained initialisation}
As aforementioned, the ViViT model~\cite{arnab2021vivit} that we use as our backbone is initialised from a ViT image model~\cite{dosovitskiy_iclr_2021} that is trained on ImageNet-21K~\cite{deng_cvpr_2009}, like other popular video architectures~\cite{bertasius_arxiv_2021,liu2022video,yan2022multiview,zhang2021vidtr}.

In this experiment, we also consider initial pretraining on the larger Weak Textual Supervision (WTS)~\cite{stroud2020learning} dataset which consists of about 60 million weakly labelled videos scraped from the web.
Table~\ref{tab:ablation_wts} shows that co-finetuning (with K400, MiT and SSv2) still improves our model's accuracy in this case by 0.9\%.
Note that our models in Tab.~\ref{tab:ablation_wts} use a larger frame resolution of 512 for the longer side, which coupled with the stronger pretraining, significantly improves the baseline model's results too.

\subsection{Comparison to the state-of-the-art}

\begin{table}[t]
	\centering
	\caption{Comparison to the state-of-the-art on AVA~\cite{gu_cvpr_2018} using the latest v2.2 annotations. Previous work followed the traditional finetuning approach of pretraining on Kinetics and then finetuning on AVA. %
	}
	\begin{tabular}{lccc}
		\toprule
		& Pretrain & Views & mAP \\ 
		\midrule
		MViT-B~\cite{fan2021multiscale}   & K400 & 1 & 27.3 \\ 
		Unified~\cite{arnab2021unified} & K400 & 6 & 27.7 \\
		WOO~\cite{chen2021watch} & K600 & 1 & 28.3 \\
		MViT-B~\cite{fan2021multiscale} & K600 & 1 & 28.7 \\
		SlowFast R101~\cite{feichtenhofer_iccv_2019} & K600 & 6 & 29.8 \\
		ACAR~\cite{pan2021actor} & K600     & 1 & 31.4       \\
		AIA~\cite{tang2020asynchronous}  & K700     &    18        & 32.3       \\
		ACAR~\cite{pan2021actor}  & K700     &       1     & 33.3       \\ 
		TubeR~\cite{zhao2021tuber} & IG$\to$K400 & 2 & 33.6 \\
		\midrule
		ViViT-B  (AVA+K400+MiT+SSv2) & -- & 1 & 31.2 \\ %
		ViViT-L  (AVA+K700+MiT+SSv2) & -- & 1 & 32.8 \\ %
		ViViT-B  (AVA+K400+MiT+SSv2) & WTS & 1 & 34.2 \\ %
		ViViT-L  (AVA+K700+MiT+SSv2) & WTS & 1 & \textbf{36.1} \\ %
		\bottomrule
	\end{tabular}
	\label{tab:sota_ava}
\end{table}
\begin{figure}[t]
    \centering
    \includegraphics[width=\textwidth]{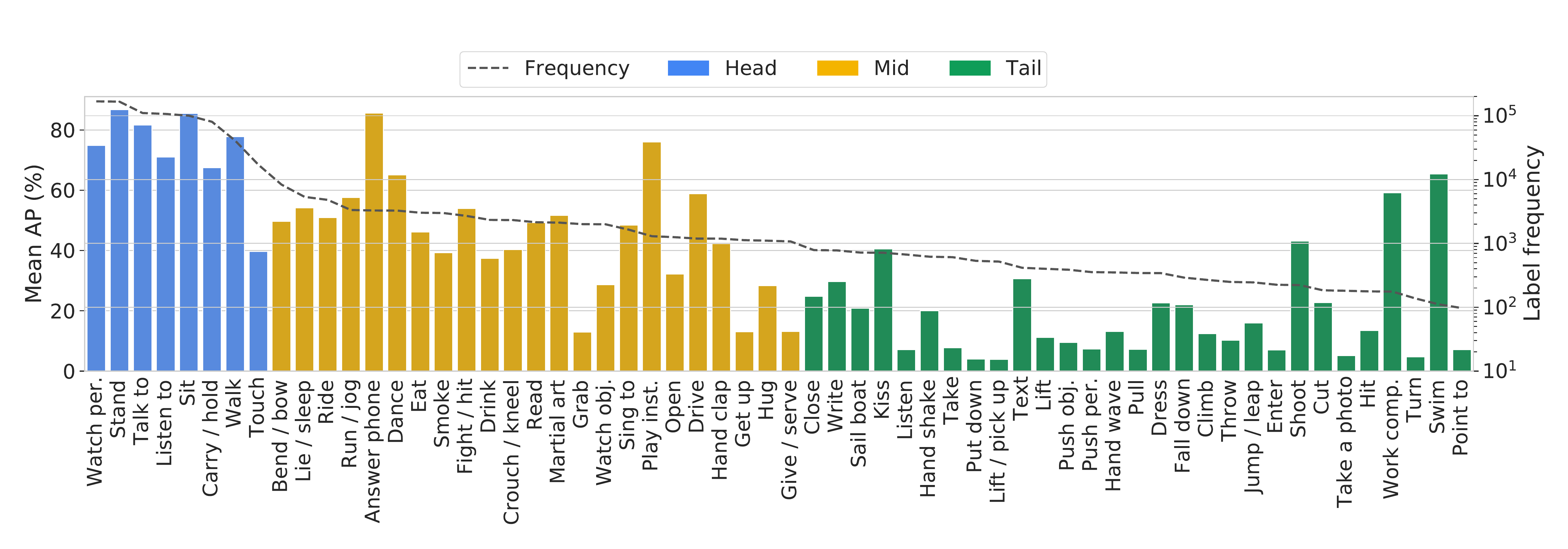}
    \caption{
    Per-class results on the AVA dataset~\cite{gu_cvpr_2018} for our best model, achieving a mean AP of 36.1.
    The label frequency, shown with a log-scale, used to divide the classes into ``Head'', ``Mid'' and ``Tail'' categories.
        }
    \label{fig:per_class_results}
\end{figure}

\subsubsection{AVA} We compare to the existing state-of-the-art approaches on AVA in Tab.~\ref{tab:sota_ava}.
All methods, excluding TubeR~\cite{zhao2021tuber} and WOO~\cite{chen2021watch} use external person detections as inputs to the model.
We use the same person detections as ~\cite{fan2021multiscale,feichtenhofer_iccv_2019,arnab2021unified,wu_cvpr_2019} as aforementioned.
For state-of-the-art comparisons, our model uses frames with a resolution of 512 on the larger side. %
Similar to object detection in images~\cite{lin2017focal,singh2018analysis}, we find that increasing the spatial resolution improves results significantly, as detailed in the supplementary.
We perform inference on a single view, %
noting that prior work often averages results over multiple resolutions, and left/right flips~\cite{feichtenhofer_iccv_2019,arnab2021unified,tang2020asynchronous,zhao2021tuber}.

Table~\ref{tab:sota_ava} shows that by co-finetuning on just publicly available classification datasets, we can achieve an AP of 32.0.
The previous highest reported result on AVA, was achieved by TubeR~\cite{zhao2021tuber} which used pretraining on the Instagram (IG) dataset~\cite{ghadiyaram2019large} containing 65 million videos.
When we pre-train our initial model with WTS~\cite{stroud2020learning} (which consists of 60 million videos, and thus of a similar scale to IG), we achieve a mean AP of 36.1 using a ViViT-Large backbone.
Figure~\ref{fig:per_class_results} shows further per-class results of our best model.

\begin{table}[t]
	\centering
	\caption{Comparison to the state-of-the-art on AVA-Kinetics~\cite{li2020ava}. All methods use a single view for inference.
	}
	\begin{tabular}{lcc}
		\toprule
		& Pretrain & mAP  \\ 
		\midrule
		Action Transformer~\cite{girdhar_cvpr_2019,li2020ava} & K400 & 23.0 \\
		ACAR~\cite{pan2021actor}  & K700              & 35.8       \\
		\midrule
		ViViT-L (AVA-Kinetics+K700+MiT+SSv2) & --  & 33.1 \\ %
		ViViT-L (AVA-Kinetics+K700+MiT+SSv2) & WTS  & \textbf{36.2} \\ %
		\bottomrule
	\end{tabular}
	\label{tab:sota_ava_kinetics}
\end{table}

\subsubsection{AVA-Kinetics}
Table~\ref{tab:sota_ava_kinetics} compares to previous methods on AVA-Kinetics.
We use the same person detections as~\cite{li2020ava}, which obtain an mAP for person detection of 77.5 on the Kinetics\textsuperscript{Box} split of this dataset.
ACAR~\cite{pan2021actor,chen20201st}, in contrast, uses a different person detector which achieves an mAP of 84.4, but has not been released.

We can separate the effect of person detections by evaluating using ground truth bounding boxes instead.
In this case, our model without WTS-pretraining achieves an mAP of 47.0 on both Kinetics\textsuperscript{Box} and the overall dataset.
ACAR is 3.4 points lower using ground truth boxes on \KINETICSBOX, attaining 43.6 (the authors do not report the overall dataset mAP in this setting).

With WTS-pretraining, our co-finetuned model achieves, to our knowledge, the best known result on this dataset with an mAP of 36.2.

\section{Conclusion and Future Work}

Recent advances in spatio-temporal action localisation have been driven by developing more complex models which use external memories to capture long-term temporal context~\cite{wu_cvpr_2019,pan2021actor,tang2020asynchronous} or construct spatio-temporal graphs of actors and objects~\cite{arnab2021unified,pan2021actor,sun_eccv_2018,zhang_tokmakov_cvpr_2019}.
In contrast, we show how we can achieve similar accuracy improvements by using a simple model, and altering the training strategy.
By leveraging our proposed co-finetuning strategy with additional, public clip-classification datasets and large-scale pretraining we were able to achieve new state-of-the-art results on the challenging AVA and AVA-Kinetics datasets. 
In particular, our co-finetuning method improved substantially on the rare classes in these long-tailed datasets, as it had a regularising effect, enabling the network to learn feature representations that transfer between different datasets.

Future work is to explore co-finetuning for other tasks and domains, and also to use more complex models.

% \clearpage
\bibliographystyle{splncs04}
\bibliography{bibliography}

\appendix

\clearpage

\begin{table}[t]
\centering
\caption{Effect of spatial resolution on performance on AVA, measured by the mAP.
We resize each frame of the video, such that the longer side of the video has the given resolution (the shorter side is zero-padded).
Observe how increasing the spatial resolution consistently improves results for both the baseline and our co-finetuning method.
Moreover, note how the improvements achieved by co-finetuning are consistent across all of these resolutions.
}
\begin{tabularx}{0.6\linewidth}{YYY}
\toprule
Resolution & Baseline & Co-finetuning \\ \midrule
224        &      23.4    &         26.3      \\
352        &     26.9     &        29.1       \\
416        &    27.5      &          30.1     \\
512        &    28.5      &          31.0     \\ \bottomrule
\end{tabularx}
\label{tab:resolution_ablation}
\end{table}

\begin{table}[b]
    \centering
    \caption{Label frequencies of each class in the AVA v2.2~\cite{gu_cvpr_2018} training set. 
    We define ``Head'' classes as having more than 10,000 labelled examples, and ``Tail'' classes having less than 1,000 examples. Classes which do not fall into either category are ``Mid'' classes.}
    \begin{tabularx}{\linewidth}{Xc|Xc|Xc}
    \toprule
    \multicolumn{6}{l}{\textit{Head} (8 classes)} \\
    \midrule
watch (a person)                         & 168148 & stand                                    & 166357 & talk to (e.g., self, a person, a group)  & 110267 \\ 
listen to (a person)                     & 106816 & sit                                      & 100323 & carry/hold (an object)                   &  80451 \\ 
walk                                     &  40771 & touch (an object)                        &  17133 & \\ 
\midrule
\multicolumn{6}{l}{\textit{Mid} (23 classes)} \\
\midrule
bend/bow (at the waist)                  &   8349 & lie/sleep                                &   5356 & ride (e.g., a bike, a car, a horse)      &   4808 \\ 
run/jog                                  &   3337 & answer phone                             &   3279 & dance                                    &   3267 \\ 
eat                                      &   3025 & smoke                                    &   2991 & fight/hit (a person)                     &   2695 \\ 
drink                                    &   2335 & crouch/kneel                             &   2321 & read                                     &   2146 \\ 
martial art                              &   2117 & grab (a person)                          &   2003 & watch (e.g., TV)                         &   1993 \\ 
sing to (e.g., self, a person, a group)  &   1643 & play musical instrument                  &   1297 & open (e.g., a window, a car door)        &   1251 \\ 
drive (e.g., a car, a truck)             &   1188 & hand clap                                &   1187 & get up                                   &   1124 \\ 
hug (a person)                           &   1103 & give/serve (an object) to (a person)     &   1073 & \\ 
\midrule
\multicolumn{6}{l}{\textit{Tail} (29 classes)} \\
\midrule
close (e.g., a door, a box)              &    786 & write                                    &    777 & sail boat                                &    719 \\ 
kiss (a person)                          &    714 & listen (e.g., to music)                  &    668 & hand shake                               &    619 \\ 
take (an object) from (a person)         &    608 & put down                                 &    533 & lift/pick up                             &    519 \\ 
text on/look at a cellphone              &    415 & lift (a person)                          &    400 & push (an object)                         &    387 \\ 
push (another person)                    &    355 & hand wave                                &    351 & pull (an object)                         &    344 \\ 
dress/put on clothing                    &    342 & fall down                                &    291 & climb (e.g., a mountain)                 &    268 \\ 
throw                                    &    249 & jump/leap                                &    245 & enter                                    &    225 \\ 
shoot                                    &    222 & cut                                      &    184 & take a photo                             &    181 \\ 
hit (an object)                          &    177 & work on a computer                       &    176 & turn (e.g., a screwdriver)               &    137 \\ 
swim                                     &    111 & point to (an object)                     &     97 &  \\
    \bottomrule
    \end{tabularx}
    \label{tab:label_distribution}
\end{table}

\section{Additional experimental details}

In this appendix, we include additional implementation details (Sec.~\ref{sec:additional_implementation_details}), analyse the effect of spatial resolution on detection performance (Sec.~\ref{sec:ablation_spatial_resolution}) and explicitly list all ``Head'', ``Mid'' and ``Tail'' classes with their label frequencies (Sec.~\ref{sec:class_frequencies}).

\subsection{Additional implementation details}
\label{sec:additional_implementation_details}

We trained all of our models with 32 frames sampled with a temporal stride of 2.
We trained our models using synchronous SGD with a momentum of 0.9, and a global batch size of 128 using TPU accelerators.

We set the base learning rate to 0.2 for all models, and followed a cosine learning rate schedule with a linear warm-up for the first 2.5 epochs.
For our baseline strategy, we finetuned all models for 20 epochs on both AVA and AVA-Kinetics.
For our co-finetuning strategy, the number of training epochs was set to 15.

For data augmentation,  we applied scale jittering to the input frames with a ratio uniformly sampled from the range $[0.65, 1.1]$.
We also applied random jittering on the person detections used during training with a maximum perturbation ratio of 0.15. 
For additional regularisation, we also applied label smoothing~\cite{szegedy_cvpr_2016} and stochastic depth~\cite{huang_stochasticdepth_eccv_2016} and their coefficients were set to 0.1 and 0.2, respectively.

For AVA, our models are trained on detected person boxes and they are first thresholded with a confidence score of 0.8 following~\cite{wu_cvpr_2019,feichtenhofer_iccv_2019,arnab2021unified,fan2021multiscale}.
For AVA-Kinetics, ground-truth person detections are used as the samples for training and we keep detections with a confidence score greater than 0.2 for evaluation.
The differences between the confidence score thresholds is because for AVA, we used the Faster-RCNN\cite{ren_neurips_2015} detector originally trained by~\cite{wu_cvpr_2019}, and for AVA-Kinetics, we used the CentreNet~\cite{zhou2019objects} detector originally trained by~\cite{li2020ava}.

\subsection{Effect of spatial resolution on performance on AVA}
\label{sec:ablation_spatial_resolution}

Table~\ref{tab:resolution_ablation} studies the effect of spatial resolution on action localisation performance on AVA.
We observe consistent improvements from increasing the spatial resolution, and this is known to improve accuracy for object detection of images too~\cite{lin2017focal,singh2018analysis}.
Moreover, the improvements achieved by co-finetuning are consistent across all the different resolutions that we evaluated.
Note that we resize each frame of the video, such that the longer side of the video has the given resolution, and the shorter side is zero-padded.

\subsection{List of head, mid and tail classes}
\label{sec:class_frequencies}

Table~\ref{tab:label_distribution} lists the ``Head'', ``Mid'' and ``Tail'' classes based on their label frequencies in the AVA v2.2~\cite{gu_cvpr_2018} training set.

We defined ``Head'' classes as having more than 10,000 labelled examples, and ``Tail'' classes have less than 1,000 examples.
Classes not falling into either of these categories are labelled ``Mid''.

\end{document}